\documentclass{article}
\PassOptionsToPackage{numbers}{natbib}
\usepackage[preprint]{neurips_2026}
\usepackage[utf8]{inputenc}
\usepackage[T1]{fontenc}
\usepackage[pagebackref=true,breaklinks=true,letterpaper=true,colorlinks,bookmarks=false]{hyperref}
\usepackage{url}
\usepackage{booktabs}
\usepackage{amsfonts}
\usepackage{nicefrac}
\usepackage{microtype}
\usepackage{enumitem}
\usepackage{amsmath}
\usepackage{bbm}
\usepackage{multirow}
\usepackage{graphicx}
\usepackage{makecell}

\title{Verification-Aware Training for Speculative Decoding}
\author{%
  Geonmo Gu$^{1,\,3}$, Byeongho Heo$^{1}$, HeeJae Jun$^{2}$, Yoohoon Kang$^{2}$, \\
  \bf Sangmin Lee$^{3}$, Sangdoo Yun$^{1}$, Dongyoon Han$^{1}$ \\
  \normalfont $^{1}$NAVER AI Lab \qquad $^{2}$NAVER AI Search Platform \qquad $^{3}$Korea University
}

\usepackage{subcaption}  
\usepackage{wrapfig}
\usepackage{colortbl}
\usepackage[cjk]{kotex}
\usepackage[dvipsnames]{xcolor}

\usepackage{lipsum}
\usepackage{graphicx}
\usepackage{multirow}
\usepackage{booktabs}
\usepackage{array}
\usepackage{amsmath,amssymb}
\usepackage{enumitem}
\usepackage{soul}
\usepackage{xspace}

\begin{document}
\maketitle

\begin{abstract}
Speculative decoding accelerates large language model inference by using a draft model to generate candidate tokens, which are verified by the target model in a single forward pass. Verification proceeds sequentially and discards every position from the first rejection onward, yet existing draft training relies on token-level imitation of the target with a fixed per-position weighting that reflects neither property. We introduce Verification-Aware Training (VAT), a plug-in framework that simulates verification at every training step and turns the resulting accept and reject patterns into supervision. VAT consists of two components: (i) a verification head, a lightweight jointly trained binary classifier that supervises the draft model on whether each position survives sequential verification; (ii) verification-adaptive weighting, which replaces the fixed weighting schedule by keeping full weight up to each sample's first rejection point and re-anchoring the decay to start there. VAT modifies only the training objective, so it can be layered on top of existing methods without changing the draft architecture, the target model, or the inference procedure. Applied to EAGLE-3 and DFlash on Qwen3-4B, Qwen3-8B, and LLaMA-3.1-8B, VAT improves average acceptance length by up to $11.4\%$ and wall-clock speedup by up to $8.7\%$, with consistent gains across math, code, and chat benchmarks. Code will be available at \href{https://github.com/naver-ai/VAT}{\texttt{https://github.com/naver-ai/VAT}}.
\end{abstract}

\section{Introduction}
\label{sec:introduction}
Large Language Models (LLMs)~\citep{brown2020language, achiam2023gpt, team2023gemini, grattafiori2024llama31, liu2024deepseek, yang2025qwen3} generate tokens autoregressively, requiring a full forward pass of the model for every generated token.
As models scale to hundreds of billions of parameters and reasoning techniques such as chain-of-thought~\citep{wei2022chain} further lengthen outputs, inference latency has become a critical bottleneck in practical deployment.
Speculative decoding~\citep{leviathan2023fast,chen2023accelerating} alleviates this bottleneck by splitting generation into two stages, drafting and verification.
A lightweight draft model first proposes several candidate tokens at low cost, and the target model then verifies all candidates in a single forward pass.
Verification proceeds sequentially from the first draft position under an acceptance rule that preserves the target distribution, and once a token is rejected, all subsequent candidates are discarded, and a new drafting round begins from the last accepted position.
Because the accepted output follows the target distribution, speculative decoding achieves \emph{lossless} acceleration and has become one of the most widely adopted inference acceleration techniques.

Recent advances have substantially pushed the frontier of speculative decoding.
EAGLE~\citep{li2024eagle} and its extensions~\citep{li2024eagle2,li2025eagle3} reuse hidden states of the target model to improve draft quality, and EAGLE-3 further improves draft training by exposing the draft model to multi-step draft-generated contexts.
More recently, DFlash~\citep{chen2026dflash} replaces autoregressive drafting with a block diffusion model~\citep{arriola2025block} that generates all draft tokens in parallel.
While differing in their drafting architectures, both directions train the draft model under the same paradigm of imitating the target model's outputs.

\begin{figure}[t]
    \centering
    \includegraphics[width=0.90\linewidth]{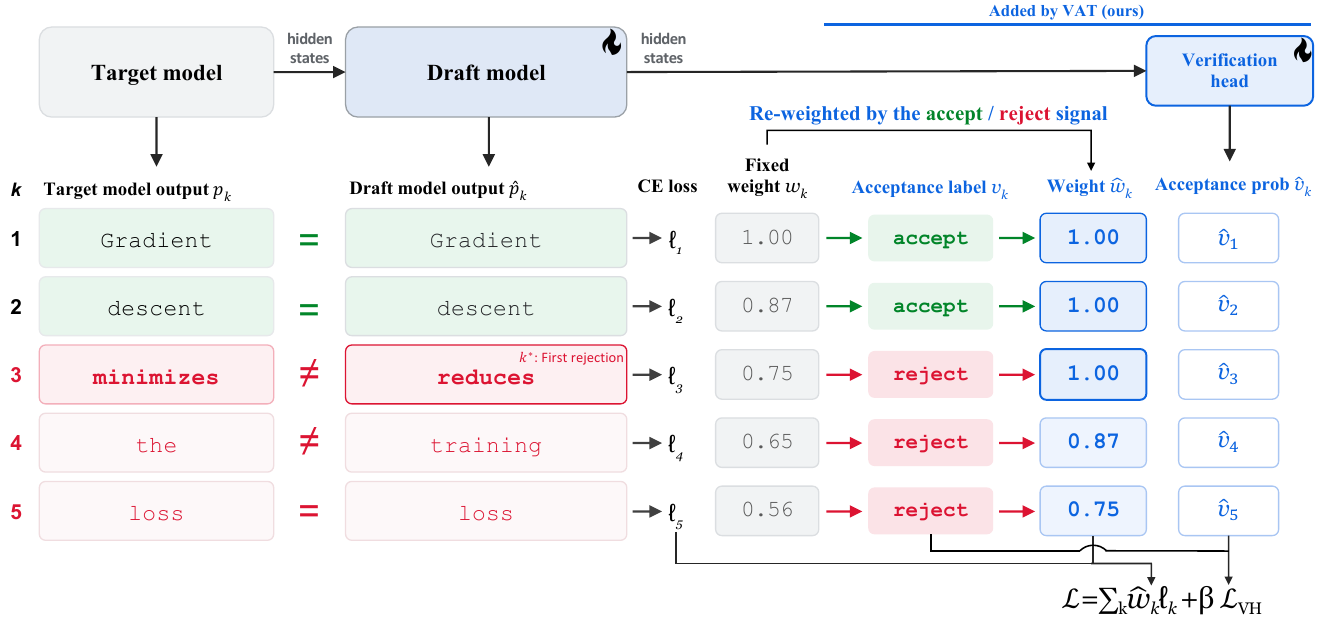}
    \caption{\textbf{Overview of Verification-Aware Training.} At each training step, VAT simulates target verification by comparing the target and draft predictions at every position $k$, yielding acceptance labels $v_k$. In this example, the first mismatch occurs at $k^* = 3$, so all subsequent positions are labeled \texttt{reject}. The labels drive two components: (i) verification-adaptive weighting, which replaces the fixed schedule $w_k$ with $\hat{w}_k$ by keeping full weight up to $k^*$ and re-anchoring the base decay to start at $k^*$, and (ii) a lightweight verification head on top of the draft's hidden states, trained to predict per-position acceptance $\hat{v}_k$. The final objective combines the reweighted cross-entropy losses with the verification head loss $\mathcal{L}_{\mathrm{VH}}$.}
    \label{fig:teaser}
    \vspace{-0.35cm}
\end{figure}

Despite these advances, the training of draft models remains misaligned with the verification process that ultimately determines speedup.
Two properties of verification are central to this misalignment.
First, the speedup of speculative decoding is governed by how many draft tokens pass the target model's verification, namely the acceptance length.
Second, verification proceeds sequentially, and once a rejection occurs at any position, all subsequent tokens are discarded regardless of their individual quality, so each draft token contributes to the acceptance length only when all of its preceding positions are accepted.
Existing training objectives, however, reflect neither of these properties.
The draft model is trained simply to imitate the target model's outputs as per-position labels, receiving no signal about whether its tokens would actually survive verification.
Although per-position loss weighting has been adopted to emphasize earlier draft positions, the schedule is fixed in advance and shared across all samples, and thus cannot adapt to where the first rejection occurs in each sample.
These observations suggest two principles for verification-aligned draft training.
First, training should go beyond imitating target outputs and incorporate verification signals from the target model.
Second, the per-position training signal should adapt to each sample's verification outcome, \ie, the position of its first rejection.

In this paper, we propose Verification-Aware Training (VAT), a simple yet effective plug-in framework that aligns draft training with the target's verification, as shown in Fig.~\ref{fig:teaser}.
VAT simulates the target's verification at every training step and realizes the two principles above through two coordinated components.
For the first principle, we propose a Verification Head, a lightweight classifier that predicts whether each draft token will be accepted by the target.
At training, we simulate verification by comparing the draft and target predictions at each position, and train the head against the resulting accept/reject labels.
Since the head is built on top of the draft's hidden states, this auxiliary objective backpropagates through the draft and shapes its representations toward features that determine agreement with the target.
This auxiliary signal complements the standard token-level prediction by reflecting verification signals.

For the second principle, we introduce Verification-Adaptive Weighting, a per-position weighting scheme that adapts to each sample's verification pattern.
As positions beyond the first rejection are discarded at inference, they should receive a reduced, though not eliminated, learning signal.
To this end, we identify the first rejection point from the simulated verification and progressively decay the weights of subsequent positions, while keeping full weight on the positions that precede it.
Since the rejection point varies across samples, the resulting schedule is sample-adaptive, unlike the fixed schedules of existing methods, and concentrates learning on the positions that determine acceptance length.

We evaluate VAT on two state-of-the-art methods spanning both drafting paradigms, the autoregressive EAGLE-3 and the diffusion-based DFlash, with three target models, Qwen3-4B, Qwen3-8B, and LLaMA-3.1-8B.
Across math, code, and chat benchmarks~\citep{cobbe2021gsm8k, lightman2023math500, chen2021humaneval, austin2021mbpp, jain2025livecodebench, zheng2023mtbench, taori2023alpaca}, VAT consistently improves both average acceptance length and wall-clock speedup on every combination of baseline and target model, improving acceptance length by up to $11.4\%$ and speedup by up to $8.7\%$.
For instance, on Qwen3-4B, VAT raises the speedup of EAGLE-3 from $4.07\times$ to $4.39\times$ (\textbf{+7.9\%}) and that of DFlash from $4.54\times$ to $4.81\times$ (\textbf{+5.9\%}), with corresponding gains in average acceptance length from $6.28$ to $6.78$ (\textbf{+8.0\%}) and from $5.73$ to $6.08$ (\textbf{+6.1\%}).
These consistent gains across target models and drafting paradigms indicate that verification-aware training yields benefits independent of the underlying drafting mechanism.

\section{Related Work}

\noindent\textbf{Speculative Decoding.}
Speculative decoding~\citep{stern2018blockwise, leviathan2023fast, chen2023accelerating, he2024rest} accelerates LLM inference by drafting candidate tokens with a lightweight model and verifying them against the target distribution in a single forward pass.
Early formulations~\citep{leviathan2023fast, chen2023accelerating} rely on a separately trained draft model from the same family as the target, which entails substantial engineering effort to obtain and align a compatible drafter~\citep{li2024eagle, cai2024medusa}.
To eliminate the dependence on a separate model, Medusa~\citep{cai2024medusa} attaches parallel decoding heads on top of the target's hidden states, and Hydra~\citep{ankner2024hydra} adds sequential dependence among these heads.
Building on this direction, EAGLE~\citep{li2024eagle} replaces parallel heads with a lightweight autoregressive drafter that operates at the feature level, with EAGLE-2~\citep{li2024eagle2} adding a dynamic draft tree and EAGLE-3~\citep{li2025eagle3} introducing multi-layer feature fusion and a training-time test that exposes the drafter to its own rollouts.
A more recent line departs from autoregressive drafting altogether and adopts diffusion-based generation~\citep{christopher2025specdiff, li2025diffuspec, sandler2025specdiff2}, producing all draft tokens in a single forward pass.
DFlash~\citep{chen2026dflash} represents the current frontier of this paradigm, training a block diffusion drafter conditioned on hidden states injected from the target.
Across both autoregressive and diffusion-based directions, draft model design has been the principal axis of innovation, while the training objective has remained largely standardized to token-level cross-entropy with sample-agnostic positional weighting.

\noindent\textbf{Training Objectives for Draft Models.}
Beyond architectural design, a smaller line of work studies the training objective for the draft model.
Knowledge distillation between draft and target distributions has been examined as a means to improve drafting quality~\citep{zhou2024distillspec, gui2024fspad, liu2024onlinesd, samarin2026lk}.
To address the train-test mismatch introduced by multi-step drafting, HASS~\citep{zhang2025hass} and the training-time test of EAGLE-3~\citep{li2025eagle3} expose the drafter to its own rollouts during training, although both retain a uniform per-position loss.
More directly relevant to the verification step, several recent works incorporate accept/reject information into the drafter, either as an inference-time signal for adaptive draft length~\citep{huang2025specdecpp, bachmann2025judge, garipov2025autojudge} or as a training-time signal that down-weights or masks the loss after rejection~\citep{bhansali2025dvi, hu2025griffin}, whereas VAT retains a decayed learning signal beyond the first rejection while concentrating learning on the prefix that determines acceptance length.
Concurrent to our work, PARD-2~\citep{an2026pard2} and D-PACE~\citep{wu2026dpace} also replace the fixed positional weighting with adaptive per-position weights, each within its own parallel drafting framework.
Specifically, PARD-2 reweights each position by the target's cumulative confidence over the preceding prefix, while D-PACE derives per-position weights from a differentiable surrogate of expected acceptance length based on the draft's own confidences.
In contrast, VAT conditions the weighting on the observed first-rejection position rather than a confidence-based proxy, couples it with a verification head supervising the cumulative acceptance outcome, and applies to both the autoregressive EAGLE-3 and the diffusion-based DFlash.

\section{Preliminaries}
\label{sec:preliminaries}

\noindent\textbf{Drafting and Verification.} Speculative decoding~\citep{leviathan2023fast,chen2023accelerating} accelerates LLM inference by alternating between \emph{drafting} and \emph{verification}. Given a prefix, a draft model $\mathcal{M}_d$ takes hidden states extracted from the target model $\mathcal{M}_t$ at the prefix positions as conditioning input and generates $K$ candidate tokens, which $\mathcal{M}_t$ evaluates in a single forward pass. Verification proceeds sequentially from the first candidate under an acceptance rule that preserves the target distribution~\citep{leviathan2023fast}. Once the first rejection occurs, all subsequent candidates are discarded regardless of their individual quality, and a new drafting cycle begins from the last accepted position. The expected number of tokens accepted per verification cycle, referred to as the average acceptance length $\tau$, is the key performance metric of a speculative decoding system. Since each verification cycle incurs a fixed cost, a larger $\tau$ produces more tokens per cycle and results in higher speedup.

\noindent\textbf{Draft Model Training.} The draft model is trained to predict $K$ future tokens at each step, with each position supervised against the target model. Two recent state-of-the-art methods, the autoregressive EAGLE-3~\citep{li2025eagle3} and the diffusion-based DFlash~\citep{chen2026dflash}, differ in their drafting mechanisms but share a common form of training objective, which can be written as:
\begin{equation}
\label{eq:standard_loss}
    \mathcal{L}_{\text{draft}} = \sum_{k=1}^{K} w_k \, \ell_k,
\end{equation}
where $\ell_k$ is the cross-entropy loss between the draft model's prediction and the target model's output at position $k$, and $w_k$ is a per-position loss weight. For $\ell_k$, EAGLE-3 uses the target's output distribution as a soft label, while DFlash uses the target's sampled token as a hard label. For $w_k$, both methods adopt a predetermined position-dependent schedule, with EAGLE-3 using $w_k = 0.8^{k-1}$ and DFlash using $w_k = \exp\!\left(-(k-1)/\gamma\right)$, giving larger weight to earlier positions under the intuition that earlier tokens are more likely to be accepted.

Neither element of this objective reflects the verification process described above. Regarding $\ell_k$, the draft model is trained simply to imitate the target model's outputs as per-position labels, receiving no signal about whether each token would actually survive sequential verification, where the contribution of position $k$ is conditioned on the acceptance of all preceding positions. Regarding $w_k$, the schedule is fixed in advance and shared across all samples, so it cannot adapt to where the first rejection occurs in each sample and continues to discount positions that would contribute fully whenever the rejection occurs late. In the next section, we introduce VAT, which addresses the former by supervising simulated verification outcomes through a verification head and the latter by anchoring the per-position weights to each sample's first rejection point.

\section{Method: Verification-Aware Training}
\label{sec:method}

This section introduces Verification-Aware Training (VAT), a simple method that can be integrated with any speculative decoding methods. First, \S\ref{sec:vat_framework} introduces the concept of verification-aware training. We present a lightweight head to the draft model and jointly train it with it (\S\ref{sec:verification_head}). Finally, \S\ref{sec:veri_weighting} and \S\ref{sec:veri_loss} describe our weighting and loss designs based on Eq.~\eqref{eq:standard_loss}. VAT simulates target verification at training time and uses the resulting accept/reject patterns as additional supervision.

\subsection{Verification at Training Time}
\label{sec:vat_framework}
During training, the draft and target models produce distributions $\hat{p}_k$ and $p_k$ at every draft position $k$, which allows VAT to simulate the verification step of speculative decoding at every training step. Following speculative sampling~\citep{leviathan2023fast, chen2023accelerating}, a token $x$ drafted at position $k$ is accepted by the target with probability $\min\!\left(1, p_k(x)/\hat{p}_k(x)\right)$, and we define the per-position acceptance indicator as
\begin{equation}
\label{eq:match}
    m_k = \mathbbm{1}\!\left[\text{the drafted token at position } k \text{ is accepted}\right].
\end{equation}
Following the sequential acceptance described in \S\ref{sec:preliminaries}, the first rejection point is
\begin{equation}
\label{eq:first_rejection}
    k^* = \min\{k : m_k = 0\},
\end{equation}
with the convention $k^* = K + 1$ when all positions are accepted. The acceptance label at position $k$ is then
\begin{equation}
\label{eq:verification}
    v_k = \mathbbm{1}[k < k^*] = \prod_{j \leq k} m_j,
\end{equation}
which is one only if every preceding position is accepted. Unlike $m_k$, which is determined at each position independently, the label $v_k$ reflects the actual outcome of sequential verification at inference, where a single rejection invalidates all subsequent tokens. VAT turns these simulated outcomes into two forms of supervision. A verification head trains the draft model to predict $v_k$ directly (\S\ref{sec:verification_head}), and a verification-adaptive weighting anchors the per-position loss weights at $k^*$ (\S\ref{sec:veri_weighting}).

\subsection{Verification Head}
\label{sec:verification_head}
The first component of VAT is a verification head, a lightweight binary classifier attached on top of the draft model's last hidden states and trained jointly with the draft model. The head serves purely as a training-time auxiliary objective, leaving the drafting and verification procedures at inference unchanged without any additional cost.

\noindent\textbf{Why Is the Verification Head Necessary?} The training objective in Eq.~\eqref{eq:standard_loss} pushes the draft model to match the target's predictive distribution at each position, but the signal that actually determines the acceptance length is the acceptance label $v_k$, which depends on the outcomes of all preceding positions. Under the training objective, the draft model receives no gradient that distinguishes two qualitatively different cases at position $k$: (1) a position whose preceding tokens have all been accepted, so that a correct prediction extends the accepted prefix, and (2) a position whose prefix has already been invalidated by an earlier rejection, so that its prediction cannot affect the acceptance length. The draft model is thus trained through local next-token supervision alone, without any signal about whether its predictions would survive sequential verification. The verification head bridges this gap by supervising the draft model directly on $v_k$.

\noindent\textbf{Design and Training.} The head is a single dense layer that maps each position's hidden state to a predicted acceptance probability $\hat{v}_k$, and is trained with binary cross-entropy against the acceptance labels:
\begin{equation}
\label{eq:accept_loss}
    \mathcal{L}_{\text{VH}} = -\frac{1}{K} \sum_{k=1}^{K} \left[ v_k \log \hat{v}_k + (1 - v_k) \log (1 - \hat{v}_k) \right].
\end{equation}
Since $v_k$ assigns zero to every position beyond the first rejection, this supervision matches the conditions under which acceptance is decided at inference. Because the head is trained jointly with the draft model, gradients from $\mathcal{L}_{\text{VH}}$ flow back through the draft model and shape its hidden states toward features that determine agreement with the target, an aspect that the training objective in Eq.~\eqref{eq:standard_loss} leaves implicit.
Although the head is not used at inference in our main experiments, its predicted $\hat{v}_k$ can additionally support early-exit drafting, which we explore in \S\ref{sec:ablation}.

\subsection{Verification-Adaptive Weighting}
\label{sec:veri_weighting}
The second component of VAT revisits the per-position weight $w_k$ in Eq.~\eqref{eq:standard_loss}. To allocate the learning signal in line with each sample's verification outcome, we replace the predetermined schedule with an instance-adaptive schedule conditioned on $k^*$:
\begin{equation}
\label{eq:dynamic_decay}
    \hat{w}_k = 
    \begin{cases} 
    1 & \text{if } k < k^*, \\ 
    w_{k - k^* + 1} & \text{if } k \geq k^*, 
    \end{cases}
\end{equation}
where $w_k$ denotes the base method's predetermined schedule (\eg, $w_k = 0.8^{k-1}$ for EAGLE-3 and $w_k = \exp(-(k-1)/\gamma)$ for DFlash). Positions before the first rejection receive full weight, since these tokens contribute to the acceptance length on this sample. Beyond $k^*$, the same decay curve is reused but shifted to start at $k^*$ instead of $k=1$, so the decay applies only to positions that follow the first rejection. As $w_1 = 1$ for both base schedules, $k^*$ itself also receives full weight.

This design is motivated by aligning the decay anchor with each sample's verification pattern. Verification discards every position after the first rejection, so position $k$'s contribution to acceptance length is conditional on all earlier positions being accepted. For a given sample, this contribution remains full up to $k^*$ and only drops beyond it, meaning the natural anchor for the decay is $k^*$ rather than $k=1$. Because $k^*$ varies across samples, this anchor is sample-specific. The predetermined schedule instead decays from $k=1$ regardless of $k^*$, which discounts positions that still contribute fully and misaligns the decay curve from where each sample's contribution actually breaks down. The first rejection position itself also receives full weight, since it is the nearest correctable failure: verification fails exactly there, and an improved prediction at this position directly extends the accepted prefix. We validate this design against alternative weighting schemes in Appendix~\ref{app:weighting_ablation}.

\subsection{Training Objective}
\label{sec:veri_loss}
We use both soft-label and hard-label cross-entropy losses as the per-position training objectives with the dynamic per-position weights $\hat{w}_k$ in Eq.~\ref{eq:dynamic_decay}. This follows the standard objective in knowledge distillation~\citep{hinton2015distilling}, combining the dense signal of the soft label with direct supervision of the hard label. Empirically, this combination yields longer acceptance length than either loss alone as shown in Table~\ref{tab:ablation_ah_dynamic}.
The full objective combines the reweighted draft loss with the verification head loss,
\begin{equation}
\label{eq:total_loss}
    \mathcal{L} = \sum_{k=1}^{K} \hat{w}_k \left( \ell_k^{\text{soft}} + \ell_k^{\text{hard}} \right) + \beta \, \mathcal{L}_{\text{VH}},
\end{equation}
where $\ell_k^{\text{soft}}$ is the cross-entropy against the target's output distribution, $\ell_k^{\text{hard}}$ is the cross-entropy against the target's sampled token, and $\beta > 0$ balances the two terms. In this work, we set $\beta$ as 1.0 for all experiments. VAT thus modifies only the training-side objective, leaving the draft architecture, the target model, and the inference procedure unchanged, so it can be layered on top of existing speculative decoding methods.

\section{Experiments}
\label{sec:experiments}

\subsection{Experimental Setup}
\label{sec:setup}
We evaluate VAT on top of EAGLE-3~\citep{li2025eagle3} and DFlash~\citep{chen2026dflash} using three target models: Qwen3 (4B, 8B)~\citep{yang2025qwen3}, and LLaMA-3.1-Instruct-8B~\citep{grattafiori2024llama31}.
For each baseline, we follow the training hyperparameters reported in the original papers~\citep{li2025eagle3, chen2026dflash}. For example, we set $\gamma=7$ for the DFlash weighting schedule.
We evaluate on three task categories: math (GSM8K~\citep{cobbe2021gsm8k}, MATH-500~\citep{lightman2023math500}, AIME25), code (HumanEval~\citep{chen2021humaneval}, MBPP~\citep{austin2021mbpp}, LiveCodeBench (LCB)~\citep{jain2025livecodebench}), and chat (MT-Bench~\citep{zheng2023mtbench}, Alpaca~\citep{taori2023alpaca}).
Following prior works~\citep{li2025eagle3, chen2026dflash}, we construct a training set by pairing Perfectblend~\citep{xu2024perfect} user prompts with responses generated by the target model for better target alignment, and train each draft model for $3$ epochs on this dataset. The responses are generated with greedy decoding, and verification during training is simulated under the same decoding scheme. Results with a training set generated at temperature $=1$ are provided in Appendix~\ref{app:corpus_temperature}. For fair comparison, both the baselines and their VAT counterparts are trained on this same corpus.
For evaluation, we report the average acceptance length $\tau$ and the decoding speedup over the autoregressive baseline.
All training and evaluation runs use NVIDIA A100 80GB GPUs with bf16 precision. We measure speedup and acceptance length $\tau$ with Hugging Face's Transformers library using the public evaluation codes\footnote{EAGLE-3: https://github.com/SafeAILab/EAGLE, DFlash: https://github.com/z-lab/dflash}.

\subsection{Main Results}
\label{sec:main_results}

Table~\ref{tab:main} reports decoding speedup and average acceptance length on Qwen3-4B, Qwen3-8B, and LLaMA-3.1-8B across math, code, and chat benchmarks. VAT improves both metrics on top of EAGLE-3 and DFlash across all three target models. The gains are most apparent in the average acceptance length $\tau$, which directly reflects how many tokens the draft contributes per verification cycle. Averaged over the eight benchmarks, EAGLE-3 + VAT improves $\tau$ by $8.0\%$ on Qwen3-4B, $5.7\%$ on Qwen3-8B, and $2.5\%$ on LLaMA-3.1-8B, and DFlash + VAT improves $\tau$ by $6.1\%$, $11.4\%$, and $3.8\%$ on the same three models. Wall-clock speedup follows the same trend, with average gains of $7.9\%$, $5.0\%$, and $3.8\%$ over EAGLE-3, and $5.9\%$, $8.7\%$, and $3.4\%$ over DFlash. A similar pattern holds at decoding temperature $=1$, with consistent improvements in both $\tau$ and speedup across all three target models for both baselines. The gains hold across three target models, two model families, and both drafting paradigms, indicating that VAT is not tied to a specific target or speculative decoding methods.

\begin{table}[t]
\centering
\caption{\textbf{Decoding speedup over the autoregressive baseline and average acceptance length ($\tau$)}, evaluated with up to 2048 generated tokens. Temperature $=0$ and $=1$ denote the decoding temperature used at evaluation. All results are measured on A100 GPUs using the HuggingFace Transformers library. Boldface indicates the best result in each setting.}
\label{tab:main}
\scriptsize
\setlength{\tabcolsep}{0.15em}
\resizebox{\linewidth}{!}{%
\begin{tabular}{llcccccc cccccc cccc cc}
\toprule
 & & \multicolumn{6}{c}{\textsc{Math}} & \multicolumn{6}{c}{\textsc{Code}} & \multicolumn{4}{c}{\textsc{Chat}} & & \\
\cmidrule(lr){3-8} \cmidrule(lr){9-14} \cmidrule(lr){15-18}
\multirow{2}{*}{Model} & \multirow{2}{*}{Method} & \multicolumn{2}{c}{GSM8K} & \multicolumn{2}{c}{MATH-500} & \multicolumn{2}{c}{AIME25} & \multicolumn{2}{c}{HumanEval} & \multicolumn{2}{c}{MBPP} & \multicolumn{2}{c}{LCB} & \multicolumn{2}{c}{MT-Bench} & \multicolumn{2}{c}{Alpaca} & \multicolumn{2}{c}{\textit{Avg.}} \\
\cmidrule(lr){3-4} \cmidrule(lr){5-6} \cmidrule(lr){7-8} \cmidrule(lr){9-10} \cmidrule(lr){11-12} \cmidrule(lr){13-14} \cmidrule(lr){15-16} \cmidrule(lr){17-18} \cmidrule(lr){19-20}
 & & Speedup & $\tau$ & Speedup & $\tau$ & Speedup & $\tau$ & Speedup & $\tau$ & Speedup & $\tau$ & Speedup & $\tau$ & Speedup & $\tau$ & Speedup & $\tau$ & Speedup & $\tau$ \\
\midrule
\multicolumn{20}{c}{\textit{Temperature} $= 0$} \\
\midrule
\multirow{4}{*}{Q3-4B}
 & EAGLE-3 & 4.94$\times$ & 7.62 & 5.00$\times$ & 7.55 & 4.46$\times$ & 6.79 & 4.33$\times$ & 6.67 & 3.99$\times$ & 6.18 & 3.75$\times$ & 5.91 & 3.35$\times$ & 5.20 & 2.74$\times$ & 4.28 & 4.07$\times$ & 6.28 \\
 & \ + VAT & \textbf{5.31}$\times$ & \textbf{8.25} & \textbf{5.23}$\times$ & \textbf{7.91} & \textbf{4.61}$\times$ & \textbf{7.01} & \textbf{4.83}$\times$ & \textbf{7.37} & \textbf{4.57}$\times$ & \textbf{7.12} & \textbf{4.11}$\times$ & \textbf{6.47} & \textbf{3.52}$\times$ & \textbf{5.41} & \textbf{2.95}$\times$ & \textbf{4.67} & \textbf{4.39}$\times$ & \textbf{6.78} \\
 & DFlash & 6.69$\times$ & 8.28 & 5.93$\times$ & 7.46 & 5.05$\times$ & 6.21 & 4.58$\times$ & 5.57 & 4.30$\times$ & 5.23 & 4.67$\times$ & 5.86 & 2.82$\times$ & 4.20 & 2.27$\times$ & 3.04 & 4.54$\times$ & 5.73 \\
 & \ + VAT & \textbf{7.01}$\times$ & \textbf{8.66} & \textbf{6.31}$\times$ & \textbf{7.91} & \textbf{5.19}$\times$ & \textbf{6.44} & \textbf{4.89}$\times$ & \textbf{5.97} & \textbf{4.67}$\times$ & \textbf{5.71} & \textbf{4.92}$\times$ & \textbf{6.19} & \textbf{3.03}$\times$ & \textbf{4.53} & \textbf{2.43}$\times$ & \textbf{3.26} & \textbf{4.81}$\times$ & \textbf{6.08} \\
\midrule
\multirow{4}{*}{Q3-8B}
 & EAGLE-3 & 4.95$\times$ & 7.54 & 5.07$\times$ & 7.52 & 4.42$\times$ & 6.69 & 4.30$\times$ & 6.47 & 3.90$\times$ & 5.95 & 3.66$\times$ & 5.66 & 3.31$\times$ & 5.04 & 2.67$\times$ & 4.12 & 4.04$\times$ & 6.12 \\
 & \ + VAT & \textbf{4.97}$\times$ & \textbf{7.68} & \textbf{5.14}$\times$ & \textbf{7.73} & \textbf{4.66}$\times$ & \textbf{6.98} & \textbf{4.62}$\times$ & \textbf{6.95} & \textbf{4.08}$\times$ & \textbf{6.25} & \textbf{4.02}$\times$ & \textbf{6.26} & \textbf{3.47}$\times$ & \textbf{5.23} & \textbf{2.94}$\times$ & \textbf{4.68} & \textbf{4.24}$\times$ & \textbf{6.47} \\
 & DFlash & 6.21$\times$ & 7.45 & 5.96$\times$ & 7.28 & 4.97$\times$ & 6.01 & 4.67$\times$ & 5.59 & 4.10$\times$ & 4.86 & 4.43$\times$ & 5.51 & 3.03$\times$ & 4.24 & 2.41$\times$ & 3.15 & 4.47$\times$ & 5.51 \\
 & \ + VAT & \textbf{7.03}$\times$ & \textbf{8.71} & \textbf{6.43}$\times$ & \textbf{8.14} & \textbf{5.16}$\times$ & \textbf{6.40} & \textbf{5.06}$\times$ & \textbf{6.22} & \textbf{4.62}$\times$ & \textbf{5.69} & \textbf{4.88}$\times$ & \textbf{6.27} & \textbf{3.14}$\times$ & \textbf{4.46} & \textbf{2.55}$\times$ & \textbf{3.26} & \textbf{4.86}$\times$ & \textbf{6.14} \\
\midrule
\multirow{4}{*}{L3.1-8B}
 & EAGLE-3 & 4.39$\times$ & 6.48 & 4.53$\times$ & 6.53 & 4.44$\times$ & 6.33 & 4.45$\times$ & 6.51 & 4.52$\times$ & 6.50 & 3.88$\times$ & 5.86 & 3.56$\times$ & 5.18 & 3.55$\times$ & 5.28 & 4.17$\times$ & 6.08 \\
 & \ + VAT & \textbf{4.51}$\times$ & \textbf{6.54} & \textbf{4.69}$\times$ & \textbf{6.65} & \textbf{4.59}$\times$ & \textbf{6.45} & \textbf{4.59}$\times$ & \textbf{6.59} & \textbf{4.69}$\times$ & \textbf{6.61} & \textbf{4.02}$\times$ & \textbf{6.01} & \textbf{3.79}$\times$ & \textbf{5.42} & \textbf{3.79}$\times$ & \textbf{5.57} & \textbf{4.33}$\times$ & \textbf{6.23} \\
 & DFlash & 4.75$\times$ & 6.30 & 4.78$\times$ & 6.41 & 4.45$\times$ & 6.06 & 4.84$\times$ & 6.37 & 4.63$\times$ & 6.08 & 4.33$\times$ & 5.77 & 2.75$\times$ & 4.33 & 2.13$\times$ & 3.26 & 4.08$\times$ & 5.57 \\
 & \ + VAT & \textbf{4.97}$\times$ & \textbf{6.66} & \textbf{4.99}$\times$ & \textbf{6.76} & \textbf{4.56}$\times$ & \textbf{6.19} & \textbf{4.96}$\times$ & \textbf{6.56} & \textbf{4.75}$\times$ & \textbf{6.27} & \textbf{4.43}$\times$ & \textbf{5.94} & \textbf{2.79}$\times$ & \textbf{4.48} & \textbf{2.27}$\times$ & \textbf{3.35} & \textbf{4.22}$\times$ & \textbf{5.78} \\
\midrule
\multicolumn{20}{c}{\textit{Temperature} $= 1$} \\
\midrule
\multirow{4}{*}{Q3-4B}
 & EAGLE-3 & 4.57$\times$ & 7.15 & 4.48$\times$ & 6.95 & 3.69$\times$ & 5.68 & 4.05$\times$ & 6.27 & 3.69$\times$ & 5.76 & 3.60$\times$ & 5.71 & 3.05$\times$ & 4.75 & 2.48$\times$ & 3.99 & 3.70$\times$ & 5.78 \\
 & \ + VAT & \textbf{4.95}$\times$ & \textbf{7.85} & \textbf{4.59}$\times$ & \textbf{7.20} & \textbf{3.85}$\times$ & \textbf{5.96} & \textbf{4.48}$\times$ & \textbf{6.99} & \textbf{4.31}$\times$ & \textbf{6.78} & \textbf{3.96}$\times$ & \textbf{6.41} & \textbf{3.17}$\times$ & \textbf{5.00} & \textbf{2.78}$\times$ & \textbf{4.58} & \textbf{4.01}$\times$ & \textbf{6.35} \\
 & DFlash & 5.71$\times$ & 7.20 & 4.82$\times$ & 6.33 & 3.39$\times$ & 4.29 & 4.19$\times$ & 5.19 & 3.95$\times$ & 4.88 & 4.15$\times$ & 5.28 & 2.62$\times$ & 3.87 & 2.17$\times$ & 2.89 & 3.88$\times$ & 4.99 \\
 & \ + VAT & \textbf{6.17}$\times$ & \textbf{7.72} & \textbf{5.11}$\times$ & \textbf{6.58} & \textbf{3.50}$\times$ & \textbf{4.45} & \textbf{4.48}$\times$ & \textbf{5.47} & \textbf{4.37}$\times$ & \textbf{5.37} & \textbf{4.47}$\times$ & \textbf{5.64} & \textbf{2.80}$\times$ & \textbf{4.11} & \textbf{2.30}$\times$ & \textbf{3.06} & \textbf{4.15}$\times$ & \textbf{5.30} \\
\midrule
\multirow{4}{*}{Q3-8B}
 & EAGLE-3 & 4.53$\times$ & 6.99 & 4.44$\times$ & 6.80 & 3.47$\times$ & 5.32 & 4.00$\times$ & 6.14 & 3.67$\times$ & 5.64 & 3.40$\times$ & 5.35 & 2.97$\times$ & 4.59 & 2.41$\times$ & 3.84 & 3.61$\times$ & 5.58 \\
 & \ + VAT & \textbf{4.67}$\times$ & \textbf{7.36} & \textbf{4.55}$\times$ & \textbf{7.05} & \textbf{3.74}$\times$ & \textbf{5.76} & \textbf{4.17}$\times$ & \textbf{6.51} & \textbf{3.84}$\times$ & \textbf{6.03} & \textbf{3.68}$\times$ & \textbf{5.88} & \textbf{3.15}$\times$ & \textbf{4.90} & \textbf{2.54}$\times$ & \textbf{4.09} & \textbf{3.79}$\times$ & \textbf{5.95} \\
 & DFlash & 5.10$\times$ & 6.37 & 4.64$\times$ & 5.93 & 3.39$\times$ & 4.32 & 3.98$\times$ & 4.91 & 3.63$\times$ & 4.44 & 4.05$\times$ & 5.12 & \textbf{2.72}$\times$ & \textbf{3.87} & 2.27$\times$ & 3.05 & 3.72$\times$ & 4.75 \\
 & \ + VAT & \textbf{5.84}$\times$ & \textbf{7.40} & \textbf{5.02}$\times$ & \textbf{6.49} & \textbf{3.56}$\times$ & \textbf{4.54} & \textbf{4.18}$\times$ & \textbf{5.15} & \textbf{4.05}$\times$ & \textbf{4.98} & \textbf{4.60}$\times$ & \textbf{5.92} & 2.63$\times$ & 3.76 & \textbf{2.35}$\times$ & \textbf{3.15} & \textbf{4.03}$\times$ & \textbf{5.17} \\
\midrule
\multirow{4}{*}{L3.1-8B}
 & EAGLE-3 & 3.74$\times$ & 5.60 & 3.21$\times$ & 4.71 & 2.00$\times$ & 2.93 & 4.00$\times$ & 5.97 & 3.97$\times$ & 5.91 & 3.45$\times$ & 5.32 & 2.68$\times$ & 4.00 & 3.10$\times$ & 4.69 & 3.27$\times$ & 4.89 \\
 & \ + VAT & \textbf{3.80}$\times$ & \textbf{5.79} & \textbf{3.33}$\times$ & \textbf{4.98} & \textbf{2.49}$\times$ & \textbf{3.74} & \textbf{4.03}$\times$ & \textbf{6.10} & \textbf{4.05}$\times$ & \textbf{6.09} & \textbf{3.58}$\times$ & \textbf{5.58} & \textbf{2.84}$\times$ & \textbf{4.32} & \textbf{3.21}$\times$ & \textbf{4.98} & \textbf{3.42}$\times$ & \textbf{5.20} \\
 & DFlash & \textbf{3.32}$\times$ & \textbf{4.87} & 2.15$\times$ & \textbf{3.48} & 1.23$\times$ & 1.73 & 3.65$\times$ & 4.94 & 3.54$\times$ & 4.71 & 3.03$\times$ & 4.16 & 1.78$\times$ & 2.87 & 1.83$\times$ & 2.66 & 2.57$\times$ & 3.68 \\
 & \ + VAT & 3.30$\times$ & 4.64 & \textbf{2.20}$\times$ & \textbf{3.48} & \textbf{1.38}$\times$ & \textbf{1.97} & \textbf{3.73}$\times$ & \textbf{4.96} & \textbf{3.61}$\times$ & \textbf{4.84} & \textbf{3.15}$\times$ & \textbf{4.26} & \textbf{1.85}$\times$ & \textbf{3.01} & \textbf{1.86}$\times$ & \textbf{2.69} & \textbf{2.64}$\times$ & \textbf{3.73} \\
\bottomrule
\end{tabular}%
}
\vspace{-0.15cm}
\end{table}

\subsection{Empirical Analysis}
\label{sec:ablation}
All ablation studies in this section are conducted on top of DFlash with Qwen3-4B, and evaluated with the temperature set to 0 unless otherwise stated.

\noindent\textbf{Effect of Components.} Table~\ref{tab:ablation_ah_dynamic} ablates the verification head, verification-adaptive weighting, and the joint use of soft and hard labels in Eq.~\eqref{eq:total_loss} on top of DFlash with Qwen3-4B. Each factor improves the average acceptance length on its own, raising $\tau$ from $5.73$ to $5.87$ with the verification head, $5.91$ with verification-adaptive weighting, and $5.82$ with the soft + hard labels. Since DFlash originally trains with hard labels alone, the gain from adding the soft label suggests that aligning the draft with the target's output distribution, beyond matching its generated token, further helps the draft agree with the target. Combining any two factors compounds these gains, and the verification head with verification-adaptive weighting achieves the highest speedup among the pairs ($4.76\times$), recovering most of the total gain with hard labels alone. Combining all three yields the best average $\tau$ ($6.08$) and the best speedup ($4.81\times$) and attains the best results in all three task categories. Explicit supervision through the verification head and per-position reweighting anchored at each sample's first rejection address different aspects of the verification process, and the gains compound when applied together.

\begin{table}[t]
\centering
\footnotesize
\caption{\textbf{Factor analysis}. We ablate the verification head, verification-adaptive weight, and the use of both soft and hard labels in Eq.~\eqref{eq:total_loss}, where unchecked denotes training with hard labels only. Results are averaged over math (GSM8K, MATH-500, AIME25), code (HumanEval, MBPP, LCB), and chat (MT-Bench, Alpaca). The best result in each column is in bold.}
\label{tab:ablation_ah_dynamic}
\setlength{\tabcolsep}{3.5pt}
\resizebox{\linewidth}{!}{%
\begin{tabular}{ccc@{\hspace{6pt}}cccccccc}
\toprule
\multirow{2}{*}{Verification head} & \multirow{2}{*}{\parbox{2.0cm}{\centering Verification-\\adaptive weight}} & \multirow{2}{*}{\parbox{1.6cm}{\centering Soft + hard\\labels}} & \multicolumn{2}{c}{Math} & \multicolumn{2}{c}{Code} & \multicolumn{2}{c}{Chat} & \multicolumn{2}{c}{\textit{Avg.}} \\
\cmidrule(lr){4-5} \cmidrule(lr){6-7} \cmidrule(lr){8-9} \cmidrule(lr){10-11}
 & & & Speedup & $\tau$ & Speedup & $\tau$ & Speedup & $\tau$ & Speedup & $\tau$ \\
\midrule
           &            &            & 5.89$\times$          & 7.32          & 4.52$\times$          & 5.55          & 2.55$\times$          & 3.62          & 4.54$\times$          & 5.73          \\
\checkmark &            &            & 5.99$\times$          & 7.53          & 4.57$\times$          & 5.64          & 2.63$\times$          & 3.73          & 4.62$\times$          & 5.87          \\ 
           & \checkmark &            & 5.97$\times$          & 7.46          & 4.67$\times$          & 5.82          & 2.64$\times$          & 3.72          & 4.65$\times$          & 5.91          \\ 
           &            & \checkmark & 5.93$\times$          & 7.40          & 4.54$\times$          & 5.63          & 2.63$\times$          & 3.74          & 4.58$\times$          & 5.82          \\ 
           & \checkmark & \checkmark & 6.02$\times$          & 7.55          & 4.69$\times$          & 5.90          & 2.69$\times$          & 3.87          & 4.69$\times$          & 6.05          \\
\checkmark &            & \checkmark & 6.09$\times$          & 7.63          & 4.66$\times$          & 5.77          & 2.71$\times$          & 3.88          & 4.72$\times$          & 5.99          \\
\checkmark & \checkmark &            & 6.12$\times$          & 7.61          & 4.75$\times$          & 5.84          & 2.70$\times$          & 3.79          & 4.76$\times$          & 5.99          \\
\checkmark & \checkmark & \checkmark & \textbf{6.17$\times$} & \textbf{7.67} & \textbf{4.83$\times$} & \textbf{5.96} & \textbf{2.73$\times$} & \textbf{3.90} & \textbf{4.81$\times$} & \textbf{6.08} \\
\bottomrule
\end{tabular}
}
\end{table}


\begin{table}[t]
\centering
\footnotesize
\caption{\textbf{Ablation on Verification-adaptive weighting.} We compare different per-token weighting strategies $w_k$. Results are averaged over math (GSM8K, MATH-500, AIME25), code (HumanEval, MBPP, LCB), and chat (MT-Bench, Alpaca). The best result in each column is in bold.}
\label{tab:ablation_decay}
\setlength{\tabcolsep}{3.5pt}
\begin{tabular}{cc@{\hspace{6pt}}cccccccc}
\toprule
\multirow{2}{*}{\parbox{2.0cm}{\centering Schedule}} & \multirow{2}{*}{\parbox{1.2cm}{\centering Base\\$w_k$}} & \multicolumn{2}{c}{Math} & \multicolumn{2}{c}{Code} & \multicolumn{2}{c}{Chat} & \multicolumn{2}{c}{\textit{Avg.}} \\
\cmidrule(lr){3-4} \cmidrule(lr){5-6} \cmidrule(lr){7-8} \cmidrule(lr){9-10}
 & & Speedup & $\tau$ & Speedup & $\tau$ & Speedup & $\tau$ & Speedup & $\tau$ \\
\midrule
Uniform & 1.0 & 5.81$\times$ & 7.30 & 4.45$\times$ & 5.52 & 2.52$\times$ & 3.62 & 4.48$\times$ & 5.72 \\
\midrule
\multirow{2}{*}{\parbox{2.0cm}{\centering Predefined\\weight}} & $0.8^{k-1}$~\citep{li2025eagle3} & 5.81$\times$ & 7.28 & 4.51$\times$ & 5.56 & 2.57$\times$ & 3.65 & 4.51$\times$ & 5.73 \\
 & $\exp(-(k-1)/\gamma)$~\citep{chen2026dflash} & 6.09$\times$ & 7.63 & 4.66$\times$ & 5.77 & 2.71$\times$ & 3.88 & 4.72$\times$ & 5.99 \\
\midrule
\multirow{2}{*}{\parbox{2.0cm}{\centering Verification-\\adaptive weight}} & $0.8^{k-1}$~\citep{li2025eagle3} & 6.04$\times$ & \textbf{7.70} & 4.73$\times$ & 5.94 & 2.68$\times$ & \textbf{3.90} & 4.71$\times$ & \textbf{6.09} \\
 & $\exp(-(k-1)/\gamma)$~\citep{chen2026dflash} & \textbf{6.17$\times$} & 7.67 & \textbf{4.83$\times$} & \textbf{5.96} & \textbf{2.73$\times$} & \textbf{3.90} & \textbf{4.81$\times$} & 6.08 \\
\bottomrule
\end{tabular}
\end{table}

\begin{figure}[t]
\centering
\begin{subfigure}[t]{0.48\linewidth}
    \centering
    \includegraphics[width=\linewidth]{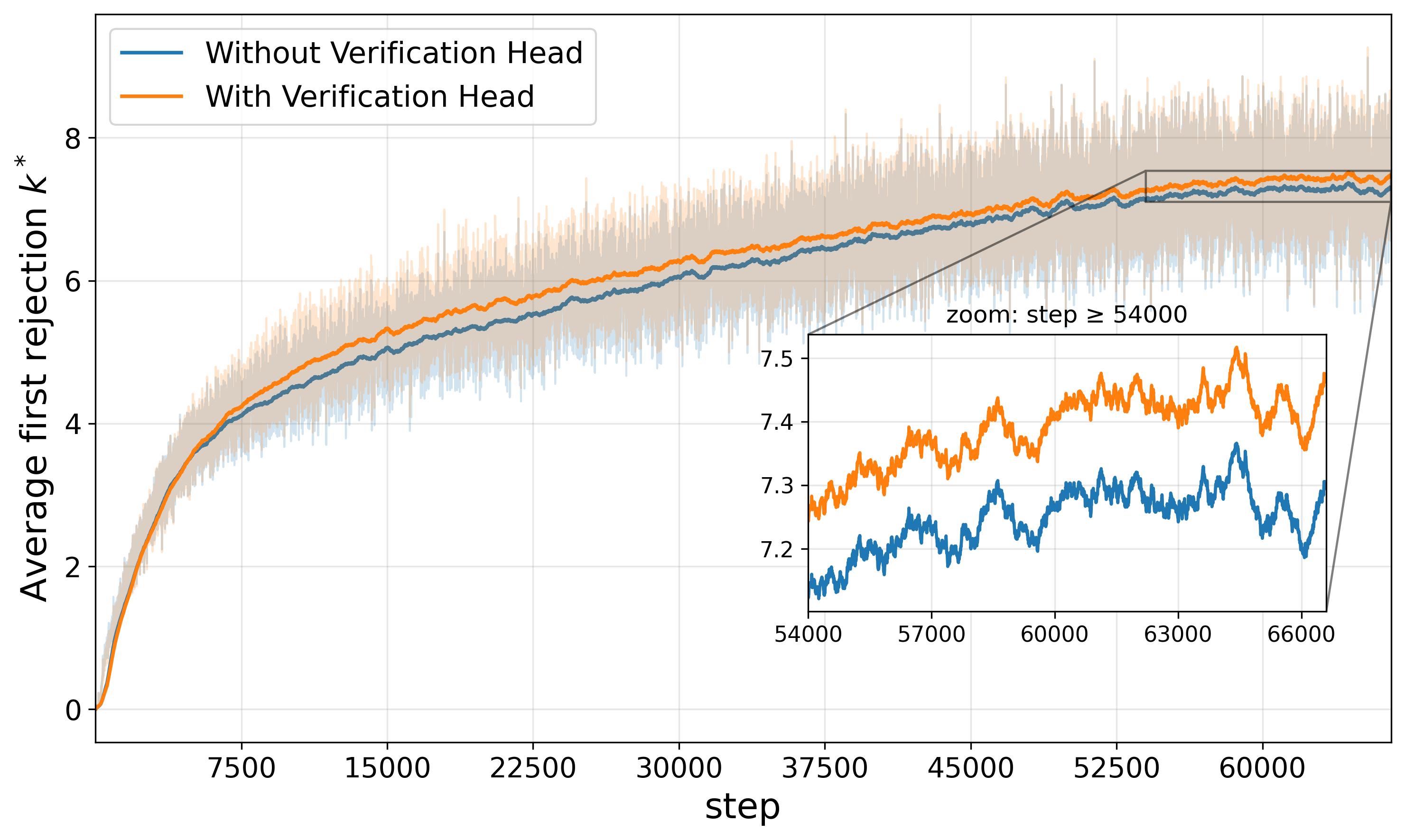}
    \caption{First rejection position}
    \label{fig:training_dynamics_a}
\end{subfigure}
\hfill
\begin{subfigure}[t]{0.48\linewidth}
    \centering
    \includegraphics[width=\linewidth]{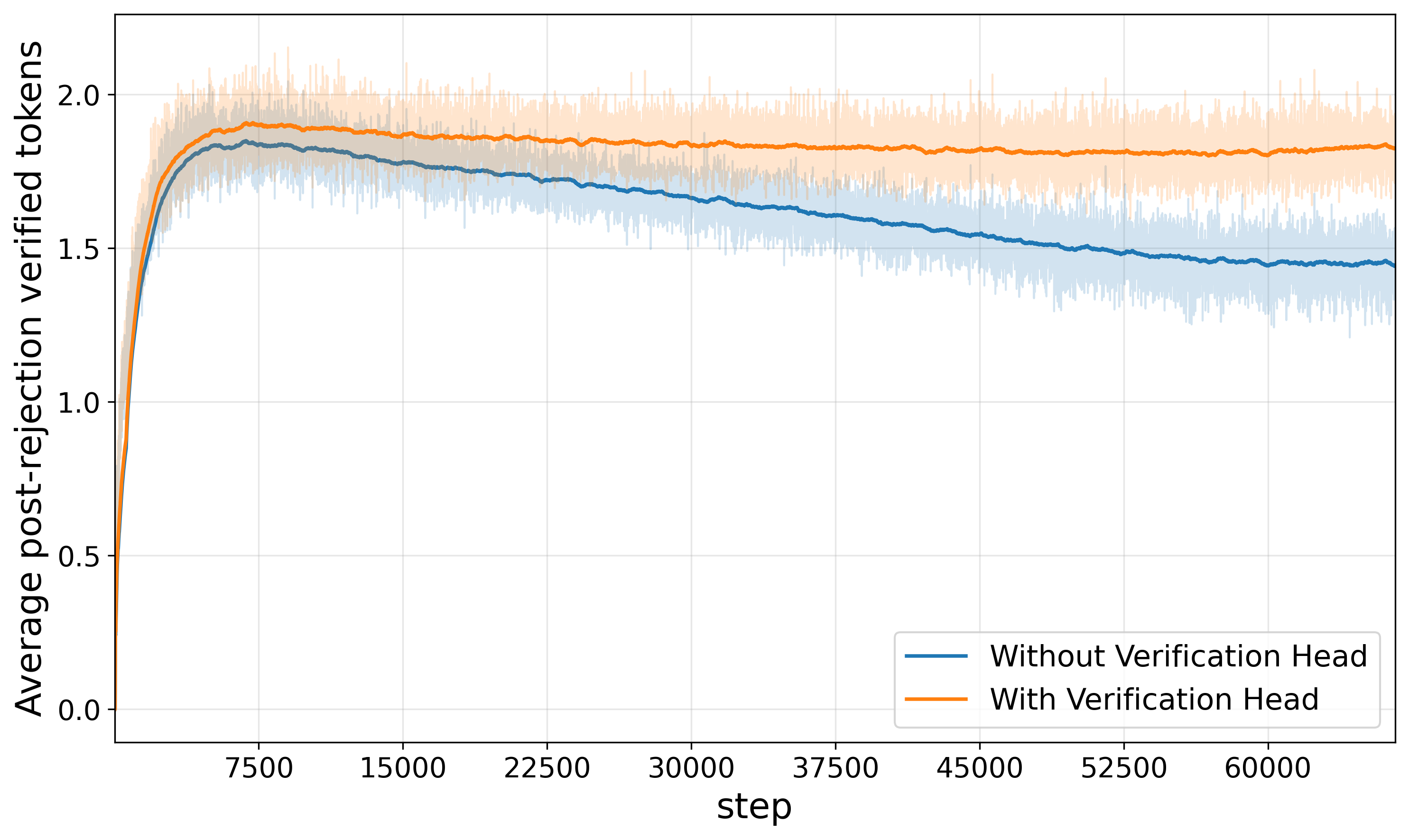}
    \caption{Post first-rejection matched tokens}
    \label{fig:training_dynamics_b}
\end{subfigure}
\caption{\textbf{Training dynamics with and without the verification head.} We track two quantities under simulated verification, averaged over the batch at each training step. (a) The first rejection position $k^*$. With the verification head, $k^*$ shifts to later positions, meaning that more consecutive draft tokens pass verification. (b) The number of post-first-rejection tokens that still match the target. The baseline gradually declines as training proceeds, while the verification-head variant remains stable, despite its post first-rejection region beginning at deeper positions on average. Curves are smoothed with a trailing moving average over a 200-step window, and the shaded thin line shows the raw per-step values.}
\label{fig:training_dynamics}
\end{figure}

\noindent\textbf{Ablation on Verification-adaptive Weighting.} Table~\ref{tab:ablation_decay} decomposes our weighting design into two factors: the choice of base weight $w_k$ and whether the weighting is made verification-adaptive. The results show two consistent patterns. First, uniform weighting and the EAGLE-3-style predefined decay $0.8^{k-1}$ produce nearly identical performance (Avg. $\tau$ of $5.72$ and $5.73$), while the DFlash-style predefined decay $\exp(-(k-1)/\gamma)$ alone improves $\tau$ to $5.99$. Second, verification-adaptive weighting achieves the best results regardless of the base weight, reaching $\tau = 6.09$ with the EAGLE-3-style base and $\tau = 6.08$ with the DFlash-style base, with a negligible difference between the two, indicating that the gain is driven primarily by the adaptation mechanism rather than by the functional form of the underlying decay. Unlike a predefined schedule that applies the same coefficient to every sample, verification-adaptive weighting concentrates the loss on positions that contribute to each sample's acceptance length. Since the rejection point varies across samples, this sample-adaptive behavior cannot be approximated by any sample-agnostic schedule.

\noindent\textbf{Training Dynamics of Verification Head.}
To examine how the verification head shapes draft behavior during training, we track two quantities across training steps in Figure~\ref{fig:training_dynamics}. The first rejection consistently occurs at later positions when the verification head is attached, as shown in Figure~\ref{fig:training_dynamics}(a). Since the head supervises whether each position survives sequential verification, its gradient flows back through the draft and steers the hidden states toward features that determine acceptance. More consecutive draft tokens therefore pass verification before the first mismatch, which directly translates into longer verified prefixes at inference time, since speedup is governed by acceptance length. A less obvious pattern emerges among post-rejection tokens, as shown in Figure~\ref{fig:training_dynamics}(b). Both variants reach a similar peak in match count early in training, after which the baseline gradually declines while the verification-head variant remains stable, even though its post-rejection tokens begin at deeper positions on average (Figure~\ref{fig:training_dynamics}(a)), where the context is harder to predict. Since the head does not alter the cross-entropy term that drives token-level predictions but instead encourages hidden states that are informative about the verification outcome, the two signals are not in conflict, and the induced representations appear to support better generalization across positions, including those past the first rejection.

\begin{figure}[t]
    \centering
    \includegraphics[width=1.0\columnwidth]{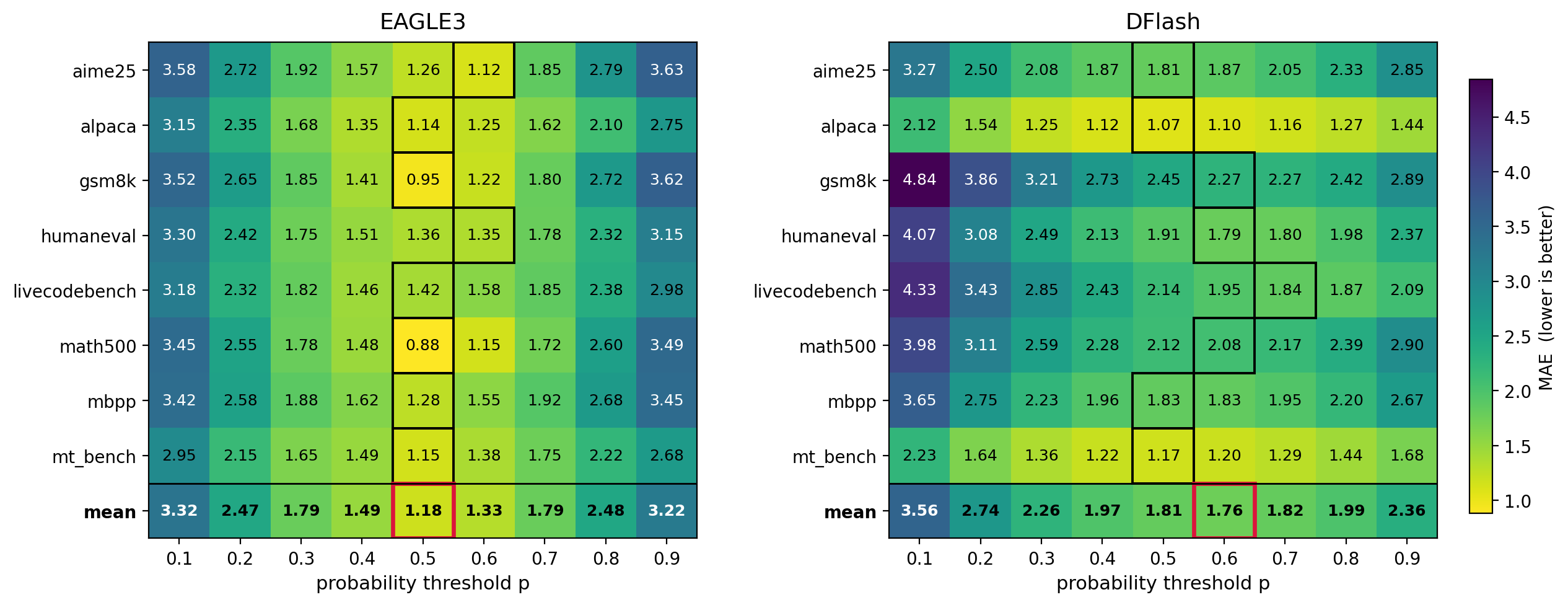}
    \caption{\textbf{First-rejection prediction error of the verification head.} Mean absolute error (MAE) between the first-rejection position obtained from target-model verification, $k^*$, and the position predicted by the verification head, as the threshold $p$ varies from $0.1$ to $0.9$ on EAGLE-3 (left) and DFlash (right). Per-token acceptance probabilities are thresholded at $p$ to locate the first predicted rejection. The bottom row reports the mean across eight benchmarks. Black outlines mark per-benchmark minima, and red outlines mark the threshold that minimizes the overall mean ($p{=}0.5$ for EAGLE-3, $p{=}0.6$ for DFlash).}
    \label{fig:head_error}
\end{figure}

\begin{figure}[t]
    \centering
    \includegraphics[width=1.0\columnwidth]{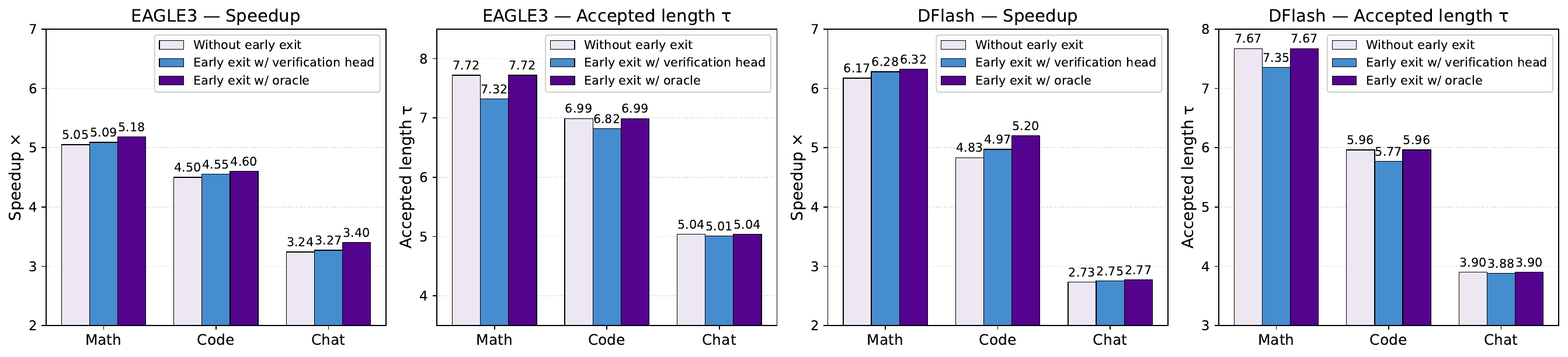}
    \caption{\textbf{Early-exit drafting guided by the verification head.} Speedup and mean acceptance length $\tau$ on Math, Code, and Chat for Qwen3-4B with VAT applied to EAGLE-3 and DFlash. ``Early exit w/ verification head'' truncates the draft at the predicted first-rejection position from the head's per-token probabilities, sending only the predicted-accept prefix to the target model. ``Early exit w/ oracle'' uses the true first-rejection position and serves as an upper bound. Early exit with the verification head improves speedup over drafting without early exit, with a small drop in $\tau$.}
    \label{fig:early_exit}
    \vspace{-0.1cm}
\end{figure}

\noindent\textbf{Verification Head at Inference.}
The verification head is trained as an auxiliary objective and is not required at inference, but it can serve as a cheap proxy for the target's verification outcome to reduce inference cost. Figure~\ref{fig:head_error} sweeps the decision threshold $p$ and reports the mean absolute error (MAE) between the head's predicted first-rejection position and the one obtained from the target. The mean MAE is minimized at $p{=}0.5$ for EAGLE-3 and $p{=}0.6$ for DFlash, with average errors of only $1.18$ and $1.76$ tokens, and we adopt these thresholds at inference. For DFlash, which drafts all $16$ tokens in a single parallel forward pass, the head sends only the predicted-accept prefix to the target. For EAGLE-3, which drafts autoregressively, the head can additionally terminate drafting once a rejection is predicted, saving both drafting and verification compute. Figure~\ref{fig:early_exit} compares early exit with the verification head against drafting without early exit and an oracle variant that uses the true first rejection point $k^{\star}$. Early exit with the verification head recovers most of the oracle speedup, and the gains are larger on DFlash (\eg, $4.83\times \to 4.97\times$ on Code, oracle $5.20\times$) than on EAGLE-3, where autoregressive drafting already dominates the compute and verification savings have less headroom. The small drop in $\tau$ (\eg, $7.67 \to 7.35$ on DFlash Math) reflects occasional false rejection predictions by the head. Overall speedup improves, since the compute saved by early exit outweighs the tokens lost to these false rejections.

\section{Conclusion}
We presented Verification-Aware Training (VAT), a plug-in framework that aligns draft model training with the target's verification process. By simulating verification at every training step, VAT introduces two components: the verification head that supervises verification outcomes as an explicit prediction target and verification-adaptive weighting that adapts per-position weights to each sample's first rejection point. Both components act only on the training objective, so VAT can be layered on top of existing speculative decoding methods without modifying the draft architecture, the target model, or the inference procedure. The verification head further admits an optional inference-time use, where its predictions guide early-exit drafting for additional speedup. Applied to EAGLE-3 and DFlash, VAT consistently improves both average acceptance length and wall-clock speedup across math, code, and chat benchmarks. Our current evaluation is limited to models up to 8B parameters. Exploring the scalability of verification-aware training on significantly larger models can be a promising future direction.

{
    \small
    \bibliographystyle{neurips_2026}
    \bibliography{main}
}

\clearpage
\newpage
\appendix
\newpage

\appendix
\setcounter{table}{0}
\setcounter{figure}{0}
\renewcommand{\thetable}{\Alph{table}}
\renewcommand{\thefigure}{\Alph{figure}}

\section{Ablation on Weighting Schemes}
\label{app:weighting_ablation}

This section validates the design of verification-adaptive weighting in Eq.~\eqref{eq:dynamic_decay} against alternative treatments of positions from the first rejection point $k^*$ onward, as well as against two schemes that reweight all positions without conditioning on $k^*$, on top of DFlash with Qwen3-4B. All variants replace only the per-position weights $\hat{w}_k$ while keeping the rest of the VAT recipe, and use a $1$-epoch training budget.

\begin{table}[h]
\vspace{-0.3cm}
\centering
\footnotesize
\caption{\textbf{Ablation on weighting schemes.} We compare verification-adaptive weighting in Eq.~\eqref{eq:dynamic_decay} against alternative treatments of positions from the first rejection point $k^*$ onward, and against two schemes that reweight all positions without conditioning on $k^*$, on top of DFlash with Qwen3-4B under a $1$-epoch training budget. Unless noted otherwise, $\hat{w}_k = 1$ for $k < k^*$. Results are averaged over math (GSM8K, MATH-500, AIME25), code (HumanEval, MBPP, LCB), and chat (MT-Bench, Alpaca). The best result in each column is in bold.}
\label{tab:app_weighting}
\setlength{\tabcolsep}{3.5pt}
\resizebox{\linewidth}{!}{%
\begin{tabular}{lc@{\hspace{6pt}}cccccccc}
\toprule
\multirow{2}{*}{Weighting scheme} & \multirow{2}{*}{\parbox{2.4cm}{\centering Weight assignment}} & \multicolumn{2}{c}{Math} & \multicolumn{2}{c}{Code} & \multicolumn{2}{c}{Chat} & \multicolumn{2}{c}{\textit{Avg.}} \\
\cmidrule(lr){3-4} \cmidrule(lr){5-6} \cmidrule(lr){7-8} \cmidrule(lr){9-10}
 & & Speedup & $\tau$ & Speedup & $\tau$ & Speedup & $\tau$ & Speedup & $\tau$ \\
\midrule
DFlash baseline & $\exp(-(k-1)/\gamma)$ & 5.51$\times$ & 7.06 & 4.25$\times$ & 5.37 & 2.45$\times$ & 3.53 & 4.27$\times$ & 5.54 \\
\midrule
Prefix-only & $0$ from $k^*$ & 3.10$\times$ & 3.94 & 2.31$\times$ & 2.89 & 1.66$\times$ & 2.21 & 2.44$\times$ & 3.11 \\
Hard cutoff & $1$ at $k^*$, then $0$ & 5.74$\times$ & 7.30 & 4.41$\times$ & 5.58 & 2.55$\times$ & 3.68 & 4.44$\times$ & 5.75 \\
Unshifted decay & $w_k$ at all $k$ & 5.79$\times$ & 7.29 & 4.43$\times$ & 5.60 & 2.52$\times$ & 3.68 & 4.46$\times$ & 5.75 \\
Marginal contribution & soft, exact derivative & 5.59$\times$ & 7.21 & 4.39$\times$ & 5.59 & 2.57$\times$ & 3.75 & 4.38$\times$ & 5.74 \\
\midrule
GRIFFIN-style masking~\citep{hu2025griffin} & $0$ on top-$m$ mismatch & 5.54$\times$ & 7.15 & 4.32$\times$ & 5.50 & 2.49$\times$ & 3.61 & 4.32$\times$ & 5.64 \\
D-PACE-style weights~\citep{wu2026dpace} & confidence-based, all $k$ & 5.77$\times$ & 7.41 & 4.53$\times$ & 5.77 & \textbf{2.62$\times$} & \textbf{3.80} & 4.52$\times$ & 5.89 \\
\midrule
Verification-adaptive weight (Eq.~\eqref{eq:dynamic_decay}) & $w_{k-k^*+1}$ from $k^*$ & \textbf{6.05$\times$} & \textbf{7.67} & \textbf{4.57$\times$} & \textbf{5.92} & 2.53$\times$ & 3.73 & \textbf{4.61$\times$} & \textbf{6.03} \\
\bottomrule
\end{tabular}
}
\end{table}

Table~\ref{tab:app_weighting} reports the results, where we refer to Eq.~\eqref{eq:dynamic_decay} as ours and cite average speedup and $\tau$ as a pair. Prefix-only assigns zero weight from $k^*$ onward and collapses below the baseline, since a sample rejected at the first draft position contributes no draft loss, and Figure~\ref{fig:training_dynamics}(a) shows $k^*$ is small in early training when the gradient is needed most. Hard cutoff differs only by keeping full weight at $k^*$ and recovers most of the gap ($4.44\times$ / $5.75$ against $4.61\times$ / $6.03$ for ours), and the residual margin indicates that positions beyond the first rejection carry useful learning signal once decayed rather than eliminated. Re-anchoring accounts for a comparable margin, as unshifted decay applies the base schedule from $k=1$ and reaches $4.46\times$ / $5.75$. Two schemes derived from the draft's own confidences land closer: the marginal-contribution scheme weights each position by the exact derivative of the expected accepted length, and the probability that verification reaches each position, which appears as a factor in this derivative, likewise discounts $k^*$ whenever the preceding prefix is uncertain ($4.38\times$ / $5.74$), while the confidence-based weights of the concurrent D-PACE~\citep{wu2026dpace} reach $4.52\times$ / $5.89$ within our framework. Masking on a per-position criterion, as in GRIFFIN~\citep{hu2025griffin}, which zeroes the loss where the drafted token falls outside the target's top-$m$ ($m{=}3$, its default), gives $4.32\times$ / $5.64$. Across all six alternatives, whether or not they condition on the first rejection, anchoring the decay at the observed $k^*$ gives the highest average speedup and $\tau$.

\section{Effect of the Verification Rule and Corpus Temperature}
\label{app:corpus_temperature}

The main experiments generate the training corpus with greedy decoding and simulate verification under the same decoding scheme (\S\ref{sec:setup}), where the acceptance rule in \S\ref{sec:vat_framework} reduces to top-1 agreement between $\hat{p}_k$ and $p_k$. This section examines whether VAT is sensitive to these choices by varying both the verification rule used during training and the temperature used to generate the corpus. All experiments use DFlash with Qwen3-4B.

\begin{table}[h]
\centering
\footnotesize
\caption{\textbf{Correlation between the two verification rules.} Pearson correlation between the accepted lengths simulated under greedy verification and under the stochastic acceptance rule, measured over training. Each column denotes the verification rule used for training.}
\label{tab:app_verification_corr}
\begin{tabular}{lcc}
\toprule
Training progress & Greedy-trained ($T{=}0$) & Sampling-trained ($T{=}1$) \\
\midrule
0--25\%   & 0.93 & 0.93 \\
25--50\%  & 0.93 & 0.93 \\
50--75\%  & 0.92 & 0.92 \\
75--100\% & 0.92 & 0.92 \\
\bottomrule
\end{tabular}
\end{table}

\begin{table}[h]
\centering
\footnotesize
\caption{\textbf{Effect of the verification rule and corpus temperature.} DFlash + VAT on Qwen3-4B trained under each combination of corpus temperature and verification rule, evaluated at both $T{=}0$ and $T{=}1$. Results are reported as speedup / $\tau$, averaged over all benchmarks.}
\label{tab:app_corpus_temp}
\begin{tabular}{llcc}
\toprule
Training corpus & Verification rule & Eval $T{=}0$ & Eval $T{=}1$ \\
\midrule
$T{=}0$ (main results) & greedy   & 4.81$\times$ / 6.08 & 4.15$\times$ / 5.30 \\
$T{=}0$                & sampling & 4.80$\times$ / 6.06 & 4.14$\times$ / 5.29 \\
$T{=}1$                & greedy   & 4.81$\times$ / 6.10 & 4.15$\times$ / 5.31 \\
$T{=}1$                & sampling & 4.83$\times$ / 6.11 & 4.17$\times$ / 5.32 \\
\bottomrule
\end{tabular}
\end{table}

Table~\ref{tab:app_verification_corr} first compares the two verification rules at the level of the training signal itself. Throughout training, the accepted lengths simulated under the two rules maintain a correlation above $0.92$ regardless of which rule is used for training, so both rules provide highly consistent information about where each draft block is first rejected. Table~\ref{tab:app_corpus_temp} then compares the downstream performance across all four combinations of corpus temperature and verification rule. The two rules differ by at most $0.02$ in both speedup and $\tau$ regardless of the corpus and evaluation temperature, and regenerating the corpus at $T{=}1$ yields nearly identical results. Consistent with the signal-level agreement above, VAT is insensitive to both the verification rule simulated during training and the corpus temperature.

\section{Training Overhead of VAT}
\label{app:training_overhead}

VAT modifies only the training objective, and this section quantifies its training-time cost. We measure the per-step training time and peak GPU memory of EAGLE-3 and DFlash with and without VAT on Qwen3-4B, averaged over $100$ steps on a single NVIDIA A100 80GB GPU. No verification labels or target distributions are precomputed or cached. All quantities are computed online at each step with the frozen target model.

\begin{table}[h]
\centering
\footnotesize
\caption{\textbf{Training overhead of VAT.} Per-step training time and peak GPU memory with Qwen3-4B, measured over $100$ steps on a single A100 80GB GPU.}
\label{tab:app_overhead}
\begin{tabular}{lcc}
\toprule
Method & Time (s/step) & Peak memory (GB) \\
\midrule
EAGLE-3 & 0.511 & 18.5 \\
+ VAT   & 0.517 & 18.6 \\
\midrule
DFlash  & 1.044 & 23.8 \\
+ VAT   & 1.108 & 31.5 \\
\bottomrule
\end{tabular}
\end{table}

Table~\ref{tab:app_overhead} reports the results. VAT adds $1.2\%$ per-step time to EAGLE-3 and $6.1\%$ to DFlash. The gap between the two baselines comes from where the target distributions required for simulating verification are obtained. EAGLE-3 already computes the target LM head distribution for its soft labels, so VAT reuses it and the remaining cost is only the lightweight verification head. DFlash never applies the target LM head in its original training, so simulating verification adds one LM head pass per step, which accounts for both the larger time increase and the higher peak memory. In both cases, the overhead applies only to training, and the inference procedure remains unchanged.


\end{document}